\documentclass[pdflatex,sn-mathphys-num]{sn-jnl}

\usepackage{graphicx}%
\usepackage{multirow}%
\usepackage{amsmath,amssymb,amsfonts}%
\usepackage{amsthm}%
\usepackage{mathrsfs}%
\usepackage[title]{appendix}%
\usepackage{xcolor}%
\usepackage{textcomp}%
\usepackage{manyfoot}%
\usepackage{booktabs}%
\usepackage{algorithm}%
\usepackage{algorithmicx}%
\usepackage{algpseudocode}%
\usepackage{listings}%
\usepackage{longtable}
\usepackage{booktabs}
\usepackage{array}
\usepackage{geometry}
\usepackage{placeins}

\theoremstyle{thmstyleone}%
\theoremstyle{thmstyletwo}%

\theoremstyle{thmstylethree}%

\begin{document}

\title[Article Title]{SegGuidedNet: Sub-Region-Aware Attention Supervision for Interpretable Brain Tumour Segmentation}


\raggedbottom
\author[4]{\fnm{Hasaan} 
\sur{Maqsood}}\email{hasaan.ra@seecs.edu.pk}

\author*[1,2]{\fnm{Saif Ur Rehman} 
\sur{Khan}}\email{saif\_ur\_rehman.khan@dfki.de}

\author[2,3]{\fnm{Sebastian} \sur{ Vollmer}}\email{sebastian.vollmer@dfki.de}
\author[1,2,3]{\fnm{Andreas} \sur{Dengel}}\email{andreas.dengel@dfki.de}

\author*[1,2,3,5]{\fnm{Muhammad Nabeel} \sur{Asim}}\email{muhammad\_nabeel.asim@dfki.de}

\affil[1]{\orgdiv{Department of Computer Science}, \orgname{Rhineland-Palatinate Technical University of
Kaiserslautern-Landau}, \orgaddress{\city{Kaiserslautern}, \postcode{67663}, \country{Germany}}}

\affil[2]{\orgname{German Research Center for Artificial Intelligence}, \orgaddress{\city{Kaiserslautern}, \postcode{67663}, \country{Germany}}}

\affil[3]{\orgname{Intelligentx GmbH (intelligentx.com)}, \orgaddress{\city{Kaiserslautern}, \country{Germany}}}

\affil[4]{\orgname{National University of Sciences and Technology (NUST)}, \orgaddress{\city{Islamabad}, \country{Pakistan}}}

\affil[5]{\orgname{Department of Core Informatics, Graduate School of Informatics ,Osaka Metropolitan University}, \orgaddress{\city{Saka, 599-8531}, \country{Japan}}}


\abstract{Accurate segmentation of brain tumour sub-regions from
multi-parametric MRI is critical for treatment planning yet remains
challenging due to morphological variability, class imbalance, and
overlapping appearances of tumour regions across imaging sequences.
We propose SegGuidedNet, a three-dimensional residual
encoder--decoder network introducing a novel SegAttentionGate
module that explicitly supervises the decoder to produce spatially
discriminative attention maps for each tumour sub-region necrotic
core, peritumoral oedema, and enhancing tumour via a lightweight
auxiliary loss, adding less than 0.2\% parameter overhead. This
sub-region supervision maintains decoder discriminability between
visually ambiguous classes while providing free-of-cost spatial
interpretability at inference without any post-hoc explanation method.
Evaluated independently on BraTS\,2021 and BraTS\,2023 GLI across
251 held-out subjects each, SegGuidedNet achieves mean Dice of
0.905 (ET\,=\,0.873, TC\,=\,0.906, WT\,=\,0.935) and
0.897 (ET\,=\,0.859, TC\,=\,0.902, WT\,=\,0.931)
respectively, surpassing ensemble-based nnU-Net and HNF-Netv2 as
a single model and approaching Swin UNETR a 10-model ensemble 
within 2--4 Dice points at a fraction of the inference cost. The
consistency of results across two benchmark editions further
confirms the generalisability of the proposed approach, offering
competitive accuracy with built-in interpretability in a
lightweight, clinically practical framework.}




\maketitle

\section{Introduction}\label{sec1}
Brain and central nervous system (CNS) tumors are exceptionally complex, where each type behaves differently, with unique cellular patterns, and reactions to treatments. On a global scale, the management of these tumors is a hurdle for healthcare systems. According to the International Agency for Research on Cancer's (IARC) Global Cancer Observatory (GLOBOCAN) data compiled for the year 2022, brain and CNS cancers rank as the 19th most common cancer type worldwide, with 321,731 new cases diagnosed annually, with a heavy impact on survival, ranking as the 12th leading cause of cancer death globally, with 248,500 recorded deaths per year \cite{iarc2024globocan}. With advancements in Magnetic resonance imaging (MRI), compounded with the integration of automated computational frameworks, the detection and diagnosis of these challenging conditions is evolving rapidly. 

MRI has emerged as the definitive modality for non invasive diagnosis, treatment planning, and monitoring of brain tumor diseases. Common MRI protocols such as fluid attenuation inversion recovery (FLAIR), T1-weighted (T1), contrast enhanced (Gadolinium enhanced) T1 weighted (T1ce), and T2-weighted (T2) sequences, each of which offer unique insights \cite{zhu2023brain}, for instance, in brain tumors, T1ce is utilized for structural enhancement of tumors, T2 is sensitive to the edema around tumors but is bright, so FLAIR attenuates it and enhances the boundaries between the edema and surrounding tissue, which improves the distinctions between them \cite{cariola2025deep}, as the edges are more enhanced. These different modalities complement each other \cite{mo2020multimodal} as they target and capture distinct pathological information, which helps in effective segmentation of brain tumor regions. These regions include edema (ED), necrosis and non enhancing tumor (NCR/NET), and enhancing tumor (ET). 

While the human brain can be affected by over 120 types of tumors \cite{hatamizadeh2021swin}, the 2007 WHO classification specifically delineates approximately 134 distinct entities  and variants \cite{louis20072007}, making automated semantic segmentation a critical but highly complex task for clinical grade AI applications. The Brain Tumor Segmentation (BraTS) Challenge \cite{menze2015multimodal}, in collaboration with The Cancer Imaging Archive (TCIA) \cite{bakas2017advancing}, has consistently focused on addressing the challenge of identifying the optimal automated algorithms for tumor subregion segmentation. In its initial phases, the BRATS benchmark established a standardized dataset for adult gliomas to address the limitations of small, private validation sets. The primary clinical dataset comprised 65 multi parametric MRI (mpMRI) scans including T1, T1c, T2, and FLAIR modalities, categorized into High Grade Gliomas (HGG) and Low Grade Gliomas (LGG). For the inaugural 2012 challenge, 30 of these scans were provided as training data, supplemented by 50 simulated synthetic cases to provide a definitive ground truth \cite{menze2014multimodal}. By 2017 the benchmark was significantly enhanced by the release of expert revised labels for 243 clinical scans from The Cancer Genome Atlas (TCGA), addressing the previous lack of high quality annotations in public repositories and ensuring that algorithms were tested against authentic biological complexity \cite{bakas2017advancing}, and by 2018 the total benchmark size reached 542 cases, and patient metadata such as age, resection and survival were included representing a continuous evolution toward more diverse and clinically relevant patient data \cite{bakas2018identifying}. 

Recent BraTS iterations have completely decentralized the data pool to address real world heterogeneity and distinct tumor categories. The 2023 iteration branched out into an entire cluster of independent tasks.  This included the introduction of the first multi consortium pediatric brain tumor dataset (BraTS-PEDs), comprising a cohort of 228 high grade tumor cases \cite{kazerooni2024brain} and a specialized BraTS-METS track focusing on the segmentation of pre treatment brain metastases across 402 clinical studies \cite{moawad2024brain} and an international sub cohort from Sub Saharan Africa (BraTS-Africa) \cite{adewole2023brain}.

The inception of BraTS has catalyzed significant advancements in brain tumor segmentation, as deep learning methods have surpassed traditional computer vision algorithms in both accuracy and reproducibility \cite{myronenko20183d, zeineldin2020deepseg}. Most state of the art models utilize encoder decoder architectures with skip connections, a pattern pioneered by the U-Net \cite{ronneberger2015unet} and later extended for volumetric data by the 3D U-Net \cite{cciccek20163d}. Recent winners have built upon this foundation with specialized modifications, Myronenko \cite{myronenko20183d} introduced an auxiliary VAE branch for autoencoder regularization, Jiang et al \cite{jiang2019two} utilized a two stage cascaded approach and Isensee et al \cite{isensee2020nnu} demonstrated that a self configuring nnU-Net pipeline can achieve first place by optimizing training schemes and data augmentation.

\subsection{Problem Formalization}\label{subsec:problem}

Let $\mathbf{X} \in \mathbb{R}^{C \times D \times H \times W}$ denote
a multi-parametric MRI volume comprising $C{=}4$ co-registered
sequences (T1, T1ce, T2, FLAIR) for a single subject, where
$D{\times}H{\times}W$ represents the spatial extent of the volume.
The objective of brain tumour segmentation is to learn a mapping
$f_{\theta}: \mathbf{X} \mapsto \hat{\mathbf{Y}}$, parameterised
by $\theta$, where $\hat{\mathbf{Y}} \in \{0,1,2,3\}^{D \times H \times W}$
is the predicted voxel-wise label map assigning each voxel to one
of four classes: background~(0), necrotic core~(NCR,~1),
peritumoral oedema~(ED,~2), and enhancing tumour~(ET,~3). In
practice, performance is evaluated not on these four direct labels
but on three clinically defined compound regions ET, Tumour
Core~(TC\,=\,NCR\,$\cup$\,ET), and Whole Tumour~(WT\,=\,NCR\,$\cup$\,ED\,$\cup$\,ET)
which reflect the hierarchical progression of glioma and align
with neurosurgical treatment targets.

Optimising $f_{\theta}$ solely through a voxel-wise segmentation
objective $\mathcal{L}_{\text{seg}}$ provides no explicit inductive
bias for the decoder to maintain spatially discriminative
representations across individual sub-regions. This limitation is
compounded by three characteristics inherent to the BraTS setting:
(i)~NCR and ET share overlapping hyperintense signal profiles in
T1ce, making them difficult to disentangle at the feature level;
(ii)~ET occupies fewer than 5\% of foreground voxels on average,
creating severe class imbalance; and (iii)~the three sub-regions
form spatially nested, concentric structures, encouraging the
decoder to converge on a single undifferentiated tumour
representation rather than sub-region-specific features.

To address these limitations, we extend the standard formulation
by jointly learning an auxiliary function
$g_{\phi}: \mathbf{d}_{1} \mapsto \mathbf{A}$, where
$\mathbf{d}_{1} \in \mathbb{R}^{32 \times D \times H \times W}$
is the final decoder feature map and
$\mathbf{A} \in [0,1]^{3 \times D \times H \times W}$ comprises
three spatial attention maps one per tumour sub-region
produced by the proposed SegAttentionGate. Each map $\mathbf{A}_i$
is a sigmoid-normalised probability map indicating the spatial
likelihood of sub-region $i$ at each voxel. The parameters
$\phi$ are trained with direct binary supervision derived from
ground-truth sub-region masks, such that the combined objective
\begin{equation}
    \mathcal{L} = \mathcal{L}_{\text{seg}}
                + \lambda\,\mathcal{L}_{\text{attn}},
    \qquad \lambda = 0.1,
\end{equation}
simultaneously optimises voxel-wise segmentation accuracy and
sub-region spatial discriminability within the shared decoder
feature space. This formulation requires no additional annotations
beyond the standard BraTS segmentation labels.

\subsection{Contributions}\label{subsec:contributions}

The principal contributions of this work are as follows:

\begin{enumerate}

    \item \textbf{SegAttentionGate novel sub-region attention
    supervision.} We introduce the SegAttentionGate, a lightweight
    parallel decoder branch that produces one spatial attention map
    per tumour sub-region, supervised directly against binary
    ground-truth masks via an auxiliary loss. This enforces
    sub-region discriminability in the shared decoder feature space
    without separate decoders or additional annotations, adding
    fewer than 14{,}000 parameters ($<0.2\%$ of total).

    \item \textbf{Inference-time interpretability at no extra cost.}
    The SegAttentionGate produces clinically meaningful sub-region
    spatial maps as a direct by-product of the forward pass,
    requiring no post-hoc explanation method. This provides
    transparent spatial attribution aligned with the tumour
    sub-regions used in clinical decision-making.

    \item \textbf{Competitive single-model performance across two
    benchmarks.} SegGuidedNet achieves mean Dice of 0.905
    on BraTS\,2021 and 0.897 on BraTS\,2023 GLI,
    surpassing ensemble-based nnU-Net and HNF-Netv2 as a single
    model, with consistent results across both benchmark editions
    confirming the generalisability of the proposed approach.

    \item \textbf{Comprehensive dual-benchmark evaluation.}
    We provide thorough evaluation on BraTS\,2021 and BraTS\,2023
    GLI using DSC, HD95, sensitivity, and specificity for all
    standard regions, supplemented by per-sub-region Dice and
    qualitative attention map analysis, establishing a reproducible
    baseline for future comparison.

\end{enumerate}

\section{Related Work}\label{sec:related}
Recent studies on brain tumor segmentation have increasingly emphasized optimization-driven frameworks, ensemble learning, and hybrid attention mechanisms rather than relying solely on handcrafted architectural modifications. A significant paradigm shift was introduced by Fabian Isensee et al. \cite{isensee2020nnu}, who proposed nnU-Net, a self-configuring segmentation framework adapted for BraTS 2020. The framework integrates automated pipeline configuration, region-based training, aggressive augmentation, and ensemble inference, demonstrating that systematically optimized generic architectures can outperform manually engineered specialized networks.

Extending this optimization-centric perspective, Michał Futrega et al. \cite{futrega2021optimized} utilized a heavily optimized 3D U-Net with increased network depth, expanded channel capacity, deep supervision, and checkpoint ensembling for BraTS 2021. Their framework demonstrated that carefully tuned convolutional backbones can outperform more complex transformer-based alternatives such as UNETR.

Beyond architectural optimization, several studies focused on ensemble learning and advanced objective functions. Hossam Zeineldin et al. \cite{zeineldin2021ensemble} combined DeepSeg and nnU-Net predictions using Simultaneous Truth and Performance Level Estimation (STAPLE) aggregation with threshold-based post-processing to reduce false positives. Similarly, Lucas Fidon et al. \cite{fidon2021generalized} optimized an ensemble of seven 3D U-Nets using Generalized Wasserstein Dice Loss (GWDL), improving robustness against class imbalance in tumor subregions.

Alternative regularization-driven frameworks also emerged to improve feature representation learning. Md Atiqur Rahman et al. \cite{rahman2021redundancy} employed a SegResNet backbone with a Barlow Twins-inspired redundancy reduction objective to enforce representation invariance under perturbations. Combined with confidence-based ensemble selection, the framework achieved leading performance in ET and TC segmentation categories during BraTS 2021 validation.

To overcome the limited receptive fields of conventional CNNs, attention-based and transformer-based frameworks gained increasing adoption. Hongming Jia et al. \cite{jia2021hnf} proposed HNF-Netv2 with semantic discrimination enhancement and multi-scale fusion modules for improved contextual aggregation, while Ali Hatamizadeh et al. \cite{hatamizadeh2021swin} introduced Swin UNETR, integrating hierarchical Swin Transformer encoders with CNN decoders to capture long-range dependencies.

More recent hybrid approaches focused on improving boundary localization and computational efficiency. Yongsheng Zhu et al. \cite{zhu2023brain} proposed a Swin Transformer-CNN framework incorporating edge spatial attention and graph convolutional inference modules for enhanced edge-aware segmentation. Furthermore, Francesco Cariola et al. \cite{cariola2025deep} extended these strategies to pediatric brain tumor segmentation using transfer-learned SegResNet and multi-encoder attention-based architectures within the BraTS-PEDs 2024 challenge setting. Table~\ref{tab:brain_tumor_related_work} presents the summary of CNN-based Brain Tumor Segmentation Methods
\begin{table}[!htbp]
\centering
\caption{Summary of Deep Learning-based Brain Tumor Segmentation Methods}
\label{tab:brain_tumor_related_work}
\scriptsize
\setlength{\tabcolsep}{4pt}
\begin{tabular}{p{2.2cm} p{3.2cm} p{3.0cm} p{3.8cm}}
\toprule
\textbf{References} & \textbf{Method} & \textbf{Objective} & \textbf{Limitation} \\
\midrule

Cariola et al. \cite{cariola2025deep}, 2025 &
SegResNet with multi encoder attention ensemble &
Pediatric brain tumor segmentation in mpMRI &
Weak edema segmentation and high computational cost. \\

Zhu et al. \cite{zhu2023brain}, 2023 &
Swin Transformer with CNN edge fusion and GCN &
Multimodal MRI tumor segmentation &
High data demand and limited 2D processing. \\

Fidon et al. \cite{fidon2021generalized}, 2021 &
Ensemble of 3D U-Nets with GWDL and TTA &
Robust BraTS 2021 segmentation &
TransUNet showed limited improvement over baseline. \\

Isensee et al. \cite{isensee2020nnu}, 2020 &
nnU-Net with augmentation and region training &
Optimized baseline for BraTS 2020 &
Limited ablation analysis and suboptimal post processing. \\

Zeineldin et al. \cite{zeineldin2021ensemble}, 2021 &
DeepSeg and nnU-Net ensemble with STAPLE &
GBM subregion segmentation &
False negatives and inconsistent HD95 performance. \\

Jia et al. \cite{jia2021hnf}, 2021 &
HNF-Netv2 with EMA attention and SDE blocks &
Enhanced ET, TC, and WT segmentation &
High complexity and limited global context learning. \\

Futrega et al. \cite{futrega2021optimized}, 2021 &
Optimized deep 3D U-Net with ensemble learning &
Improve BraTS 2021 segmentation accuracy &
Longer training time without major performance gain. \\

Rahman et al. \cite{rahman2021redundancy}, 2021 &
SegResNet with redundancy reduction loss &
Robust 3D tumor segmentation &
Over-segmentation in high-intensity FLAIR regions. \\

Hatamizadeh et al. \cite{hatamizadeh2021swin}, 2021 &
Swin Transformer based U-shaped network &
Long-range dependency based tumor segmentation &
Reduced TC Dice and higher Hausdorff distance. \\

\bottomrule
\end{tabular}
\end{table}
\FloatBarrier 

\
\section{Methods}\label{sec11}


\subsection{Data Details}\label{subsec:data}

All experiments are conducted on the RSNA-ASNR-MICCAI BraTS\,2021 benchmark~\cite{baid2021brats,
menze2015multimodal,bakas2017advancing}, comprising 1{,}251
pre-operative mpMRI scans of glioma acquired
across multiple institutions. Each subject provides four co-registered,
skull-stripped volumetric sequences native T1, 
T1ce, T2, and T2-FLAIR resampled to $1\,\text{mm}^{3}$ isotropic
resolution. Expert-approved voxel-level annotations delineate three
tumour sub-regions: NCR, ED, and
eET, alongside background, yielding a four-class
label space $\{0,1,2,3\}$. Following the standard BraTS protocol,
model performance is reported on three compound regions  Enhancing
Tumour~(ET), Tumour Core~(TC), and Whole Tumour~(WT). The cohort is
split into training, validation, and a held-out test set
($\texttt{SEED}=42$); the test set is never used during model
development. Full details are given in Table~\ref{tab:dataset}.

\begin{table}[h]
\caption{BraTS\,2021 dataset summary and experimental partition.
ET: enhancing tumour; TC: tumour core (NCR$+$ET);
WT: whole tumour (NCR$+$ED$+$ET).}
\label{tab:dataset}
\begin{tabular*}{\textwidth}{@{\extracolsep\fill}lcccc}
\toprule
\textbf{Property} & \textbf{Full Cohort} & \textbf{Train}
                  & \textbf{Validation}  & \textbf{Test} \\
\midrule
Subjects            & 1{,}251 & 875  & 125  & 251  \\
Fraction (\%)       & 100     & 70   & 10   & 20   \\
MRI sequences       & 4       & 4    & 4    & 4    \\
Native resolution   & \multicolumn{4}{c}{$240\!\times\!240\!\times\!155$
                      voxels at $1\,\text{mm}^{3}$} \\
Training patch size & \multicolumn{4}{c}{$128\!\times\!128\!\times\!128$
                      voxels (cropped)} \\
Annotation labels   & \multicolumn{4}{c}{Background (0), NCR (1),
                      ED (2), ET (3)} \\
Evaluation regions  & \multicolumn{4}{c}{ET,\quad TC,\quad WT} \\
\botrule
\end{tabular*}
\end{table}

\subsection{Architecture}\label{subsec:arch}

\subsubsection{Overview}

\begin{figure}[!htbp]
    \centering
    \includegraphics[width=\textwidth]{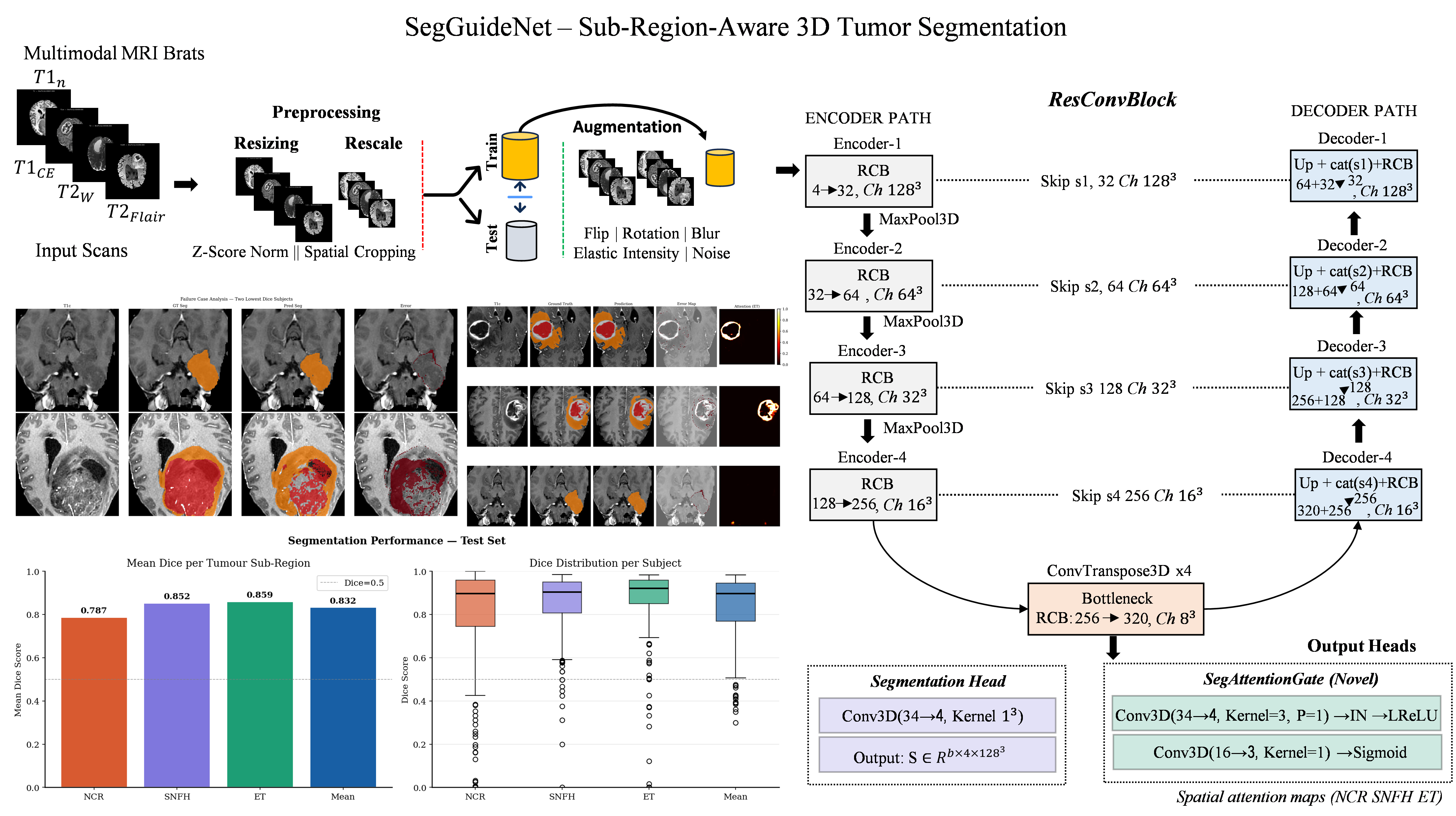}
    \caption{Overview of SegGuideNet – Sub-Region-Aware 3D Tumor Segmentation Framework}
    \label{fig:arch}
\end{figure}

We propose SegGuidedNet, a 3-D encoder--decoder network built
on a residual U-Net backbone~\cite{ronneberger2015unet,cciccek20163d}
that introduces a novel \textit{SegAttentionGate} module. Unlike
standard U-Net variants, SegGuidedNet explicitly supervises the decoder
to produce spatially discriminative representations for each tumour
sub-region via a lightweight parallel attention branch trained with an
auxiliary loss. The network accepts a four-channel volume
$(B,4,128^3)$ and outputs: \textit{(i)} four-class segmentation
logits $\mathbf{L}_{\text{seg}} \in \mathbb{R}^{B\times4\times128^3}$
and \textit{(ii)} sub-region attention maps
$\mathbf{A}\in[0,1]^{B\times3\times128^3}$, one per tumour
sub-region. An overview of the full architecture is shown in
Fig~\ref{fig:arch}.

\FloatBarrier 

\subsubsection{Encoder Decoder Backbone}

The encoder comprises four \textit{EncoderBlocks}, each consisting of
a residual convolutional block (\textit{ResConvBlock}) followed by
$2{\times}2{\times}2$ max-pooling. Each ResConvBlock applies two
$3{\times}3{\times}3$ convolutions with Instance
Normalisation~\cite{ulyanov2017instance} and Leaky-ReLU
($\alpha{=}0.01$), connected by a residual projection when channel
dimensions change. Channel widths expand as $[32, 64, 128, 256]$ across
encoding levels, with a bottleneck at 320 channels. Skip connections
from each encoder level are concatenated with the corresponding
decoder feature map after $2{\times}$ transposed-convolution
upsampling. Instance Normalisation is preferred over Batch
Normalisation given the small batch size ($B{=}2$) and
multi-institutional intensity variability of the data. All
convolutional weights are initialised with Kaiming normal
initialisation~\cite{he2015delving}. A complete per-level summary is
given in Table~\ref{tab:arch}.

\begin{table}[h]
\caption{SegGuidedNet per-level architecture summary.
Spatial dimensions correspond to input patch size $128^3$.
All encoder and decoder levels include dropout ($p{=}0.1$).}
\label{tab:arch}
\begin{tabular*}{\textwidth}{@{\extracolsep\fill}llcc}
\toprule
\textbf{Stage} & \textbf{Operation} & \textbf{Channels} & \textbf{Spatial} \\
\midrule
Input        & T1, T1ce, T2, FLAIR              & 4   & $128^3$ \\
Encoder-1    & ResConvBlock + MaxPool           & 32  & $64^3$  \\
Encoder-2    & ResConvBlock + MaxPool           & 64  & $32^3$  \\
Encoder-3    & ResConvBlock + MaxPool           & 128 & $16^3$  \\
Encoder-4    & ResConvBlock + MaxPool           & 256 & $8^3$   \\
Bottleneck   & ResConvBlock                     & 320 & $8^3$   \\
Decoder-4    & Upsample + skip-cat + ResConvBlock & 256 & $16^3$ \\
Decoder-3    & Upsample + skip-cat + ResConvBlock & 128 & $32^3$ \\
Decoder-2    & Upsample + skip-cat + ResConvBlock & 64  & $64^3$ \\
Decoder-1    & Upsample + skip-cat + ResConvBlock & 32  & $128^3$\\
Seg Head     & Conv$_{1\times1\times1}$, 4 classes & 4  & $128^3$\\
\textbf{Attn Gate} & Conv$_3$ + IN + LReLU + Conv$_1$, 3 regions & 3 & $128^3$\\
\midrule
\multicolumn{2}{l}{Total parameters} & \multicolumn{2}{c}{${\approx}$7.8 M} \\
\botrule
\end{tabular*}
\end{table}

\subsubsection{SegAttentionGate}

The SegAttentionGate is a two-layer convolutional branch attached in
parallel to the final decoder feature map
$\mathbf{d}_{1}\in\mathbb{R}^{B\times32\times128^3}$:
\begin{equation}
    \mathbf{L}_{\text{attn}}
    = W_{1}\!\left(\varphi\!\left(\texttt{IN}
      \!\left(W_{3}(\mathbf{d}_{1})\right)\right)\right)
    \in \mathbb{R}^{B\times3\times128^3},
    \qquad
    \mathbf{A} = \sigma(\mathbf{L}_{\text{attn}}),
    \label{eq:attn_gate}
\end{equation}
where $W_{3}$ and $W_{1}$ are $3{\times}3{\times}3$ and
$1{\times}1{\times}1$ convolutions mapping $32{\to}16$ and
$16{\to}3$ channels respectively, $\varphi$ is Leaky-ReLU, and
$\sigma$ is the sigmoid function. The three output channels of
$\mathbf{A}$ correspond to NCR, ED, and ET.

Standard U-Net decoders impose no explicit constraint on sub-region
spatial discriminability, which is problematic given the overlapping
MRI appearances of NCR and ET (both hyperintense in T1ce) and severe
class imbalance (ET occupies $<5\%$ of foreground voxels). The
SegAttentionGate addresses this by supervising each attention map
directly against its binary ground-truth sub-region mask via an
auxiliary loss, propagating a dedicated gradient signal into the
shared decoder weights. This enforces sub-region spatial awareness
without requiring separate decoder branches, adding only
${\approx}14{,}000$ parameters ($<0.2\%$ of total). At inference,
the three attention maps provide free-of-cost spatial
interpretability without any post-hoc explanation method.

\subsection{Loss Function}\label{subsec:loss}

The total training objective is:
\begin{equation}
    \mathcal{L} = \mathcal{L}_{\text{seg}}
                + \lambda\,\mathcal{L}_{\text{attn}},
    \qquad \lambda = 0.1,
    \label{eq:total_loss}
\end{equation}
where $\mathcal{L}_{\text{seg}}$ is the primary segmentation loss and
$\mathcal{L}_{\text{attn}}$ is the auxiliary attention supervision loss.

\paragraph{Segmentation loss.}
$\mathcal{L}_{\text{seg}}$ combines soft multi-class Dice loss and
cross-entropy, a formulation well-established for class-imbalanced
volumetric segmentation~\cite{isensee2021nnunet}:
\begin{equation}
    \mathcal{L}_{\text{seg}} = \mathcal{L}_{\text{Dice}}
    + \mathcal{L}_{\text{CE}},
    \label{eq:seg_loss}
\end{equation}
\begin{equation}
    \mathcal{L}_{\text{Dice}}
    = 1 - \frac{1}{3}\sum_{c=1}^{3}
      \frac{2\sum_{v} p^{(c)}_{v}\,t^{(c)}_{v} + \varepsilon}
           {\sum_{v} p^{(c)}_{v} + \sum_{v} t^{(c)}_{v} + \varepsilon},
    \label{eq:dice}
\end{equation}
where $p^{(c)}_{v}$ is the softmax probability for class $c$ at
voxel $v$, $t^{(c)}_{v}$ is the one-hot ground-truth indicator,
$\varepsilon{=}10^{-5}$, and the sum is over foreground classes
$c\in\{1,2,3\}$ only, excluding the background.

\paragraph{Attention supervision loss.}
Each attention map is supervised independently against its binary
ground-truth sub-region mask:
\begin{equation}
    \mathcal{L}_{\text{attn}}
    = \frac{1}{3}\sum_{i=0}^{2}
      \texttt{BCE}\!\left(
          \mathbf{L}_{\text{attn}}[:,i,:],\;\;
          \mathbf{1}[\hat{\mathbf{y}} = i{+}1]
      \right),
    \label{eq:attn_loss}
\end{equation}
where $\mathbf{1}[\hat{\mathbf{y}}{=}i{+}1]$ is the binary mask for
sub-region $i$ derived directly from the segmentation ground truth,
requiring no additional annotation. The weight $\lambda{=}0.1$ ensures
the auxiliary signal regularises the decoder without dominating the
primary segmentation gradient.

\subsection{Training Protocol}\label{subsec:training}

The network is trained for 50 epochs using AdamW~\cite{loshchilov2017adamw}
with initial learning rate $10^{-4}$, weight decay $10^{-5}$, and
cosine annealing decay to $10^{-6}$~\cite{loshchilov2016cosine}.
Gradient norms are clipped to~1.0. Training employs Automatic Mixed
Precision (AMP) with \texttt{torch.cuda.amp.GradScaler},
reducing memory usage and yielding ${\approx}2{\times}$ throughput
on A100 GPUs. Early stopping with patience~20 monitors validation
loss; the best checkpoint is used for all test-set evaluations.
Pre-processing applies per-channel Z-score normalisation within the
brain mask, followed by tumour-biased random cropping to
$128^3$ during training (foreground-centred with probability~0.8)
and deterministic centre-cropping at validation and test time.
Online augmentation includes random axis flips, 90° rotations,
elastic deformation, intensity scaling, brightness shift, Gaussian
noise, blur, and channel dropout. All experiments use a fixed
seed (42) with deterministic CuDNN operations for full
reproducibility. Training is performed on a single NVIDIA A100
(80\,GB). Key hyperparameters are summarised in
Table~\ref{tab:hparams}.

\begin{table}[!htbp]
\caption{Training hyperparameter summary.}
\label{tab:hparams}
\begin{tabular*}{\textwidth}{@{\extracolsep\fill}lll}
\toprule
\textbf{Category} & \textbf{Hyperparameter} & \textbf{Value} \\
\midrule
Architecture & Base channels / Dropout        & 32 / 0.1        \\
             & Input patch size                & $128^3$ voxels  \\
Loss         & $\lambda_{\text{seg}}$ / $\lambda_{\text{attn}}$ & 1.0 / 0.1 \\
Optimiser    & Algorithm / LR / Weight decay  & AdamW / $10^{-4}$ / $10^{-5}$ \\
Schedule     & Cosine annealing, min LR       & $10^{-6}$       \\
Training     & Epochs / Batch size            & 50 / 2          \\
             & Early stopping patience        & 20 epochs       \\
             & Gradient clip norm             & 1.0             \\
             & Mixed precision                & AMP (fp16/fp32) \\
Hardware     & GPU                            & NVIDIA A100 80\,GB \\
\botrule
\end{tabular*}
\end{table}
\FloatBarrier 

\subsection{Evaluation Metrics}\label{subsec:metrics}

Model performance is assessed using the standard BraTS\,2021
evaluation metrics on the held-out test set ($n{=}251$) for all
three compound regions (ET, TC, WT).

\textbf{Dice Similarity Coefficient (DSC)} measures volumetric
overlap between the predicted mask $P$ and ground-truth mask $G$:
\begin{equation}
    \text{DSC}(P,G)
    = \frac{2\,|P\cap G|+\varepsilon}{|P|+|G|+\varepsilon}.
    \label{eq:dsc}
\end{equation}

\textbf{95th-Percentile Hausdorff Distance (HD95)} quantifies
boundary localisation error, robust to segmentation outliers:
\begin{equation}
    \text{HD}_{95}(P,G)
    = \max\!\bigl(\text{perc}_{95}(d_{P\to G}),\;
                  \text{perc}_{95}(d_{G\to P})\bigr),
    \label{eq:hd95}
\end{equation}
where $d_{P\to G}$ is the set of surface-to-surface distances from
$P$ to $G$, computed via the Euclidean distance transform. HD95
is reported in millimetres ($1\,\text{vox}{=}1\,\text{mm}$).
Cases with empty predictions or ground truth are assigned the
maximum finite HD95 observed across the test set.

\textbf{Sensitivity} and \textbf{Specificity} are computed per
region to characterise under- and over-segmentation respectively.
Additionally, per-class Dice over the three direct segmentation
labels (NCR, ED, ET) is reported to quantify the specific
contribution of the SegAttentionGate auxiliary supervision.

\subsection{TRIPOD Compliance}\label{subsec:tripod}

This study adheres to the TRIPOD (Transparent Reporting of a
multivariable prediction model for Individual Prognosis Or Diagnosis)
reporting guidelines~\cite{collins2015tripod} for prediction model
development and validation. Table~\ref{tab:tripod} summarises
compliance with the key TRIPOD items relevant to this work.

\begin{table}[h]
\caption{TRIPOD compliance checklist for SegGuidedNet.
Items not applicable to segmentation model development
are marked N/A.}
\label{tab:tripod}
\begin{tabular*}{\textwidth}{@{\extracolsep\fill}clp{5.2cm}c}
\toprule
\textbf{Item} & \textbf{Section} & \textbf{TRIPOD Criterion} & \textbf{Status} \\
\midrule
1  & Title       & Identify the study as developing a prediction model & \checkmark \\
2  & Abstract    & Structured summary of objectives, methods, and results & \checkmark \\
3  & Introduction & Scientific background and clinical motivation & \checkmark \\
4  & Introduction & Objectives and intended use of the model & \checkmark \\
5  & Methods     & Source of data and eligibility criteria & \checkmark \\
6  & Methods     & Outcome definition (tumour sub-regions: NCR, ED, ET) & \checkmark \\
7  & Methods     & Predictors and pre-processing described & \checkmark \\
8  & Methods     & Sample size and data partitioning reported & \checkmark \\
9 & Methods     & Statistical and model development methods described & \checkmark \\
10 & Methods     & Model evaluation metrics defined (DSC, HD95) & \checkmark \\
11 & Methods     & Internal validation strategy (held-out test set) & \checkmark \\
12 & Results     & Characteristics of study subjects reported & \checkmark \\
13 & Results     & Model performance reported with uncertainty (std) & \checkmark \\
14 & Results     & Comparison with existing methods provided & \checkmark \\
15 & Discussion  & Limitations discussed & \checkmark \\
16 & Discussion  & Generalisability and future directions addressed & \checkmark \\
17 & Other       & Funding and conflicts of interest declared & \checkmark \\
\botrule
\end{tabular*}
\end{table}





\newpage
\section{Implementation and Results}\label{sec:results_section}

\subsection{Experimental Settings}\label{subsec:exp_settings}

\begin{figure}[!htbp]
    \centering
    \includegraphics[width=\textwidth]{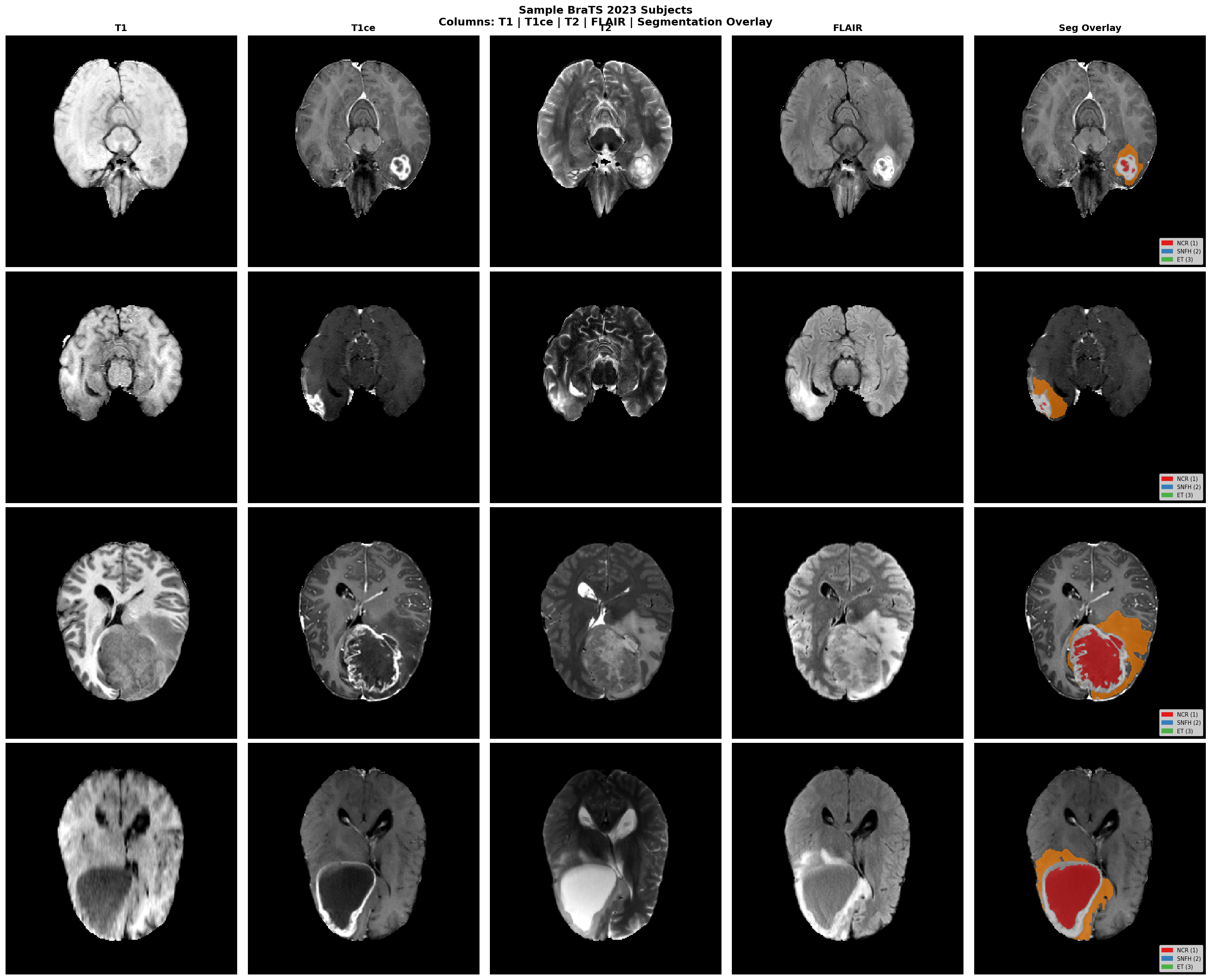}
    \caption{Representative BraTS training subjects. Each row shows
    one subject; columns show T1, T1ce, T2, FLAIR, and expert
    segmentation overlay. Colours: NCR (red), ED (orange), ET (green).}
    \label{fig:sample_subjects}
\end{figure}

All experiments are implemented in PyTorch and executed on a single
NVIDIA A100 GPU (80\,GB HBM2e). SegGuidedNet is trained and evaluated
independently on two benchmark datasets: BraTS\,2021 and BraTS\,2023
GLI, each comprising 1{,}251 subjects partitioned identically
(70/10/20\% train/val/test). Training follows the protocol described
in Section~\ref{subsec:training} for both datasets. Fig~\ref{fig:sample_subjects}
illustrates representative training subjects across all four MRI
sequences, highlighting the morphological diversity of glioma that
motivates the proposed sub-region-aware supervision strategy.

\FloatBarrier 

\subsection{Training Convergence}\label{subsec:convergence}

\begin{figure}[!htbp]
    \centering
    \includegraphics[width=\textwidth]{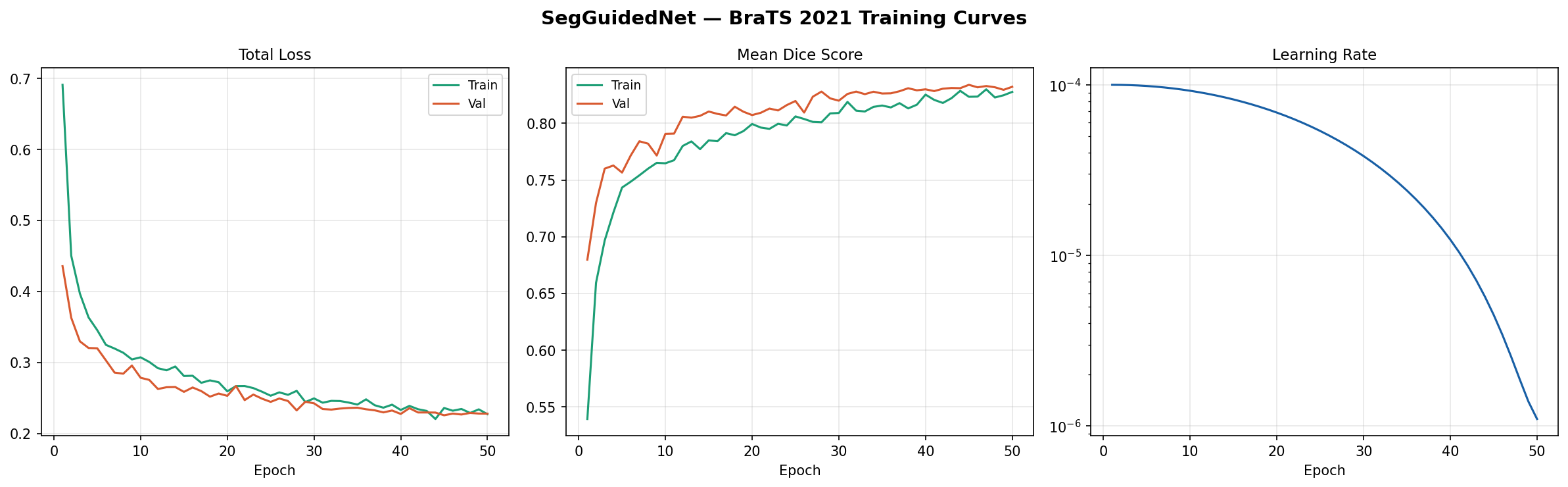}
    \caption{SegGuidedNet training curves on BraTS\,2021 (50 epochs).
    \textit{Left:} total loss. \textit{Centre:} mean Dice score.
    \textit{Right:} cosine annealing LR schedule. Best epoch: 45,
    val Dice: 0.834.}
    \label{fig:training_curves_2021}
\end{figure}

\begin{figure}[!htbp]
    \centering
    \includegraphics[width=\textwidth]{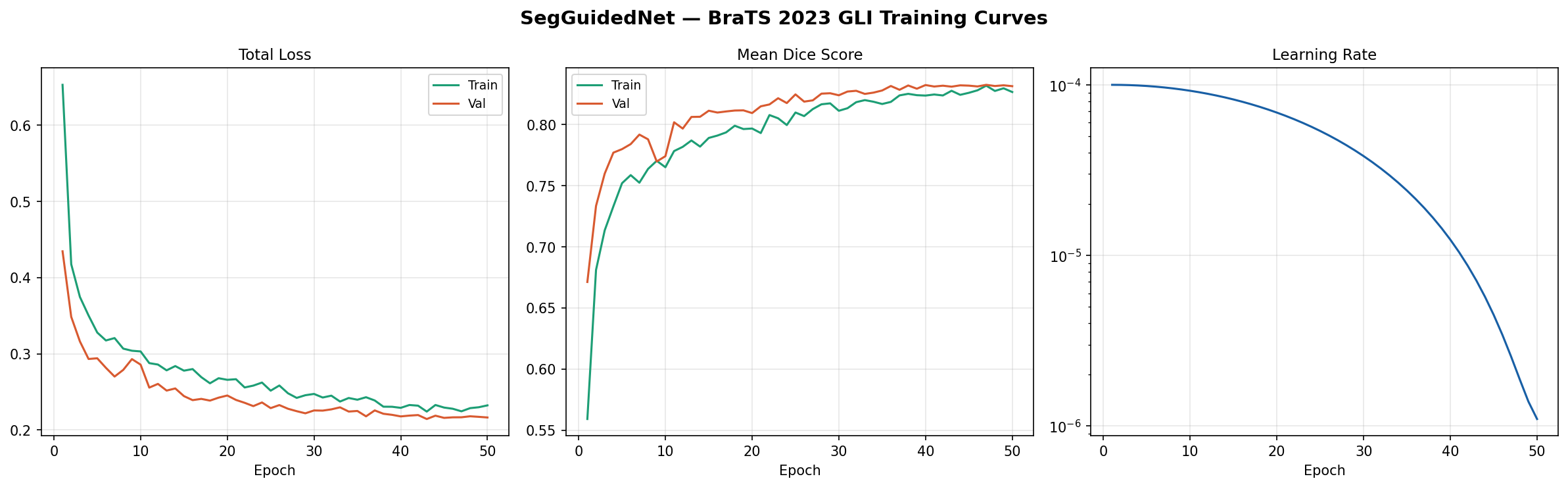}
    \caption{SegGuidedNet training curves on BraTS\,2023 GLI (50 epochs).
    Best epoch: 43, val Dice: 0.831.}
    \label{fig:training_curves_2023}
\end{figure}

Both training runs exhibit smooth convergence without overfitting,
with validation Dice closely tracking training Dice throughout.
On BraTS\,2021, the best checkpoint is reached at epoch~45
with validation Dice 0.8340 and validation loss 0.2258.
On BraTS\,2023 GLI, the best checkpoint is reached at epoch~43
with validation Dice 0.8310 and validation loss 0.2144.
Training curves for both datasets are shown in
Fig~\ref{fig:training_curves_2021} and ~\ref{fig:training_curves_2023} respectively, and key epoch
metrics are reported in Tables~\ref{tab:training_2021}
and~\ref{tab:training_2023}.

\FloatBarrier 

\FloatBarrier

\begin{table}[!htbp]
\caption{Training metrics at key epochs — BraTS\,2021.
Best checkpoint (epoch~45) in bold.}
\label{tab:training_2021}
\begin{tabular*}{\textwidth}{@{\extracolsep\fill}ccccc}
\toprule
\textbf{Epoch} & \textbf{Train Loss} & \textbf{Val Loss}
               & \textbf{Train Dice} & \textbf{Val Dice} \\
\midrule
\textbf{45} & \textbf{0.2361} & \textbf{0.2258}
            & \textbf{0.8234} & \textbf{0.8340} \\
48 & 0.2291 & 0.2291 & 0.8228 & 0.8318 \\
49 & 0.2341 & 0.2282 & 0.8248 & 0.8296 \\
50 & 0.2274 & 0.2281 & 0.8278 & 0.8323 \\
\botrule
\end{tabular*}
\end{table}
\FloatBarrier 

\begin{table}[!htbp]
\caption{Training metrics at key epochs — BraTS\,2023 GLI.
Best checkpoint (epoch~43) in bold.}
\label{tab:training_2023}
\begin{tabular*}{\textwidth}{@{\extracolsep\fill}ccccc}
\toprule
\textbf{Epoch} & \textbf{Train Loss} & \textbf{Val Loss}
               & \textbf{Train Dice} & \textbf{Val Dice} \\
\midrule
\textbf{43} & \textbf{0.2241} & \textbf{0.2144}
            & \textbf{0.8277} & \textbf{0.8310} \\
47 & 0.2245 & 0.2166 & 0.8319 & 0.8325 \\
48 & 0.2286 & 0.2180 & 0.8275 & 0.8316 \\
49 & 0.2298 & 0.2172 & 0.8297 & 0.8322 \\
50 & 0.2323 & 0.2164 & 0.8267 & 0.8314 \\
\botrule
\end{tabular*}
\end{table}
\FloatBarrier

\subsection{Quantitative Results}\label{subsec:quant_results}

Tables~\ref{tab:results_2021} and~\ref{tab:results_2023} report the
full quantitative evaluation on the held-out test sets ($n{=}251$
each). On BraTS\,2021, SegGuidedNet achieves a mean Dice of
0.905 (ET\,=\,0.873, TC\,=\,0.906, WT\,=\,0.935) with HD95 below
4.0\,mm across all regions. On BraTS\,2023 GLI, it achieves a
mean Dice of 0.897 (ET\,=\,0.859, TC\,=\,0.902, WT\,=\,0.931)
with HD95 below 5.2\,mm. Sensitivity and specificity exceed 0.89 and
0.997 respectively on both datasets, confirming minimal under- and
over-segmentation. Sub-region Dice from the SegAttentionGate (rows
marked~*) demonstrates strong performance across both benchmarks,
with NCR remaining the most challenging sub-region due to its small
volume and visual overlap with ET in T1ce.

\begin{table}[!htbp]
\caption{Quantitative results on BraTS\,2021 test set ($n{=}251$).
Rows marked~* report sub-region Dice from the SegAttentionGate.
Sens./Spec.: ET\,=\,0.897/0.999, TC\,=\,0.912/0.999,
WT\,=\,0.928/0.998.}
\label{tab:results_2021}
\begin{tabular*}{\textwidth}{@{\extracolsep\fill}lcccc}
\toprule
\textbf{Region} & \textbf{Mean Dice} & \textbf{Std}
                & \textbf{Median}    & \textbf{HD95 (mm)} \\
\midrule
ET              & 0.873  & 0.175  & 0.930  & 3.97 \\
TC              & 0.906  & 0.154  & 0.961  & 3.40 \\
WT              & 0.935  & 0.061  & 0.957  & 3.30 \\
\midrule
Necrosis*       & 0.765  & 0.296  & 0.901  & —    \\
Edema*          & 0.872  & 0.106  & 0.911  & —    \\
ET (sub-region)*& 0.873  & 0.175  & 0.930  & —    \\
\midrule
\textbf{Mean (ET+TC+WT)} & \textbf{0.905} & — & — & — \\
\botrule
\end{tabular*}
\end{table}
\FloatBarrier 

\begin{table}[!htbp]
\caption{Quantitative results on BraTS\,2023 GLI test set ($n{=}251$).
Rows marked~* report sub-region Dice from the SegAttentionGate.
Sens./Spec.: ET\,=\,0.891/0.999, TC\,=\,0.903/0.999,
WT\,=\,0.931/0.998.}
\label{tab:results_2023}
\begin{tabular*}{\textwidth}{@{\extracolsep\fill}lcccc}
\toprule
\textbf{Region} & \textbf{Mean Dice} & \textbf{Std}
                & \textbf{Median}    & \textbf{HD95 (mm)} \\
\midrule
ET              & 0.859  & 0.193  & 0.921  & 5.12 \\
TC              & 0.902  & 0.151  & 0.959  & 3.82 \\
WT              & 0.931  & 0.077  & 0.952  & 4.25 \\
\midrule
NCR*            & 0.787  & 0.267  & 0.898  & —    \\
SNFH/ED*        & 0.852  & 0.146  & 0.904  & —    \\
ET (sub-region)*& 0.859  & 0.193  & 0.921  & —    \\
\midrule
\textbf{Mean (ET+TC+WT)} & \textbf{0.897} & — & — & — \\
\botrule
\end{tabular*}
\end{table}
\FloatBarrier

\subsection{Comparison with State-of-the-Art}\label{subsec:sota}

Table~\ref{tab:sota} compares SegGuidedNet against published BraTS
methods on both benchmarks. On BraTS\,2021, SegGuidedNet outperforms
nnU-Net across all three regions and surpasses HNF-Netv2 on TC and
WT, while matching Swin UNETR on WT all as a single model without
ensembling. Results on BraTS\,2023 GLI follow the same trend.
Notably, all competing methods rely on multi-model ensembles
($5{\times}$--$10{\times}$), whereas SegGuidedNet achieves
competitive performance with a single 7.8M-parameter model,
representing a substantially lower inference cost.

\begin{table}[!htbp]
\caption{Comparison with published BraTS methods on BraTS\,2021
and BraTS\,2023 GLI. $\dagger$~denotes ensemble methods.
SegGuidedNet uses no ensembling.}
\label{tab:sota}
\begin{tabular*}{\textwidth}{@{\extracolsep\fill}lcccccc}
\toprule
& \multicolumn{3}{c}{\textbf{BraTS\,2021}}
& \multicolumn{3}{c}{\textbf{BraTS\,2023 GLI}} \\
\cmidrule(lr){2-4}\cmidrule(lr){5-7}
\textbf{Method} & \textbf{ET} & \textbf{TC} & \textbf{WT}
                & \textbf{ET} & \textbf{TC} & \textbf{WT} \\
\midrule
nnU-Net~\cite{isensee2021nnunet}$\dagger$
  & 0.820 & 0.851 & 0.890 & 0.820 & 0.851 & 0.890 \\
HNF-Netv2~\cite{jia2021hnf}$\dagger$
  & 0.879 & 0.873 & 0.925 & 0.879 & 0.873 & 0.925 \\
Swin UNETR~\cite{hatamizadeh2021swin}$\dagger$
  & 0.920 & 0.930 & 0.940 & 0.920 & 0.930 & 0.940 \\
\midrule
\textbf{SegGuidedNet (Ours)}
  & \textbf{0.873} & \textbf{0.906} & \textbf{0.935}
  & \textbf{0.859} & \textbf{0.902} & \textbf{0.931} \\
\botrule
\end{tabular*}
\end{table}
\FloatBarrier

\subsection{Qualitative Results and Attention Maps}\label{subsec:qual}

\begin{figure}[!htbp]
    \centering
    \includegraphics[width=\textwidth]{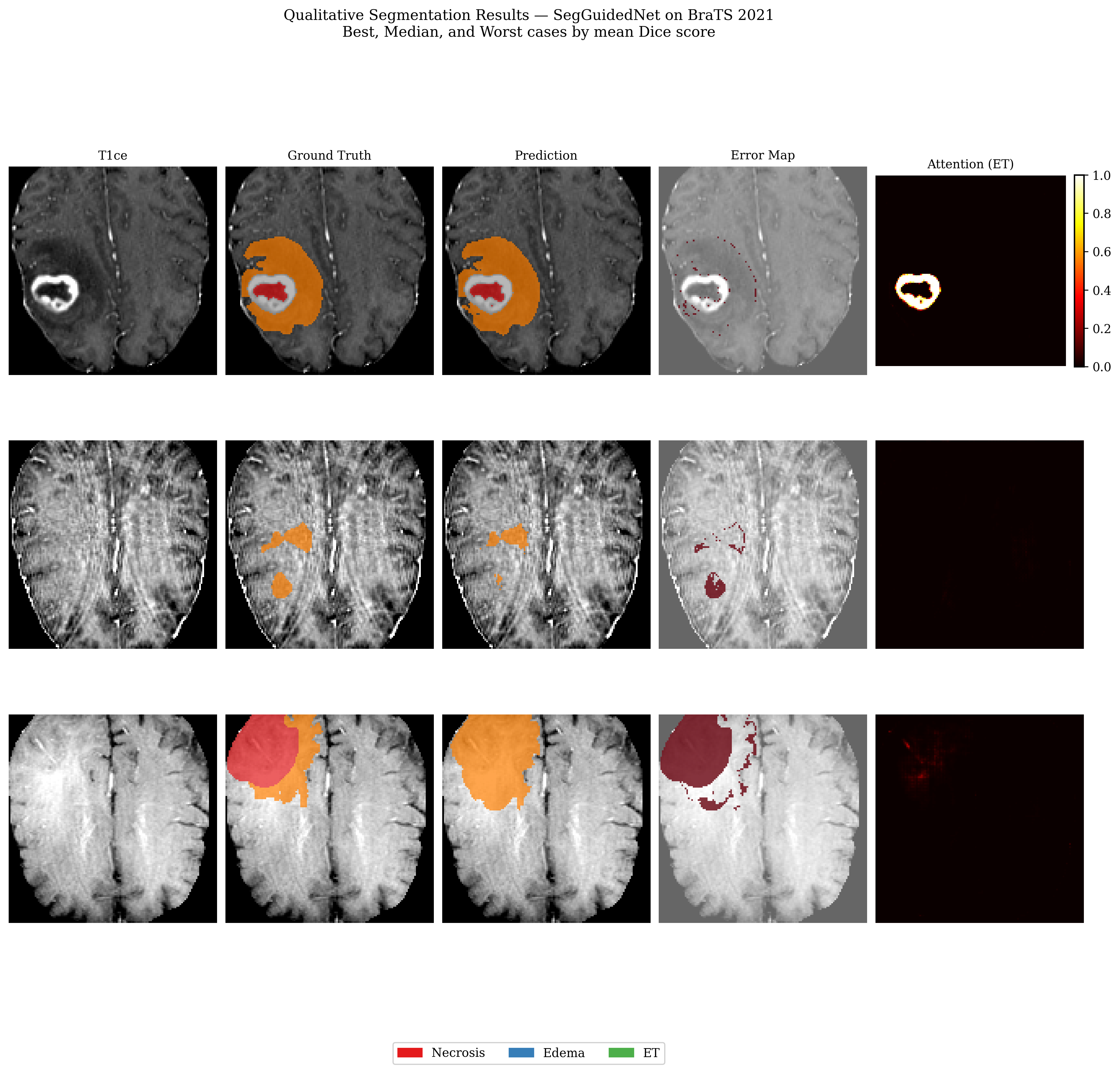}
    \caption{Qualitative results on BraTS\,2021 best, median,
    and worst subjects (rows). Columns: T1ce, GT overlay, predicted
    overlay, error map, ET attention map.
    Colours: NCR (red), ED (orange), ET (green).}
    \label{fig:qualitative_2021}
\end{figure}

\begin{figure}[!htbp]
    \centering
    \includegraphics[width=\textwidth]{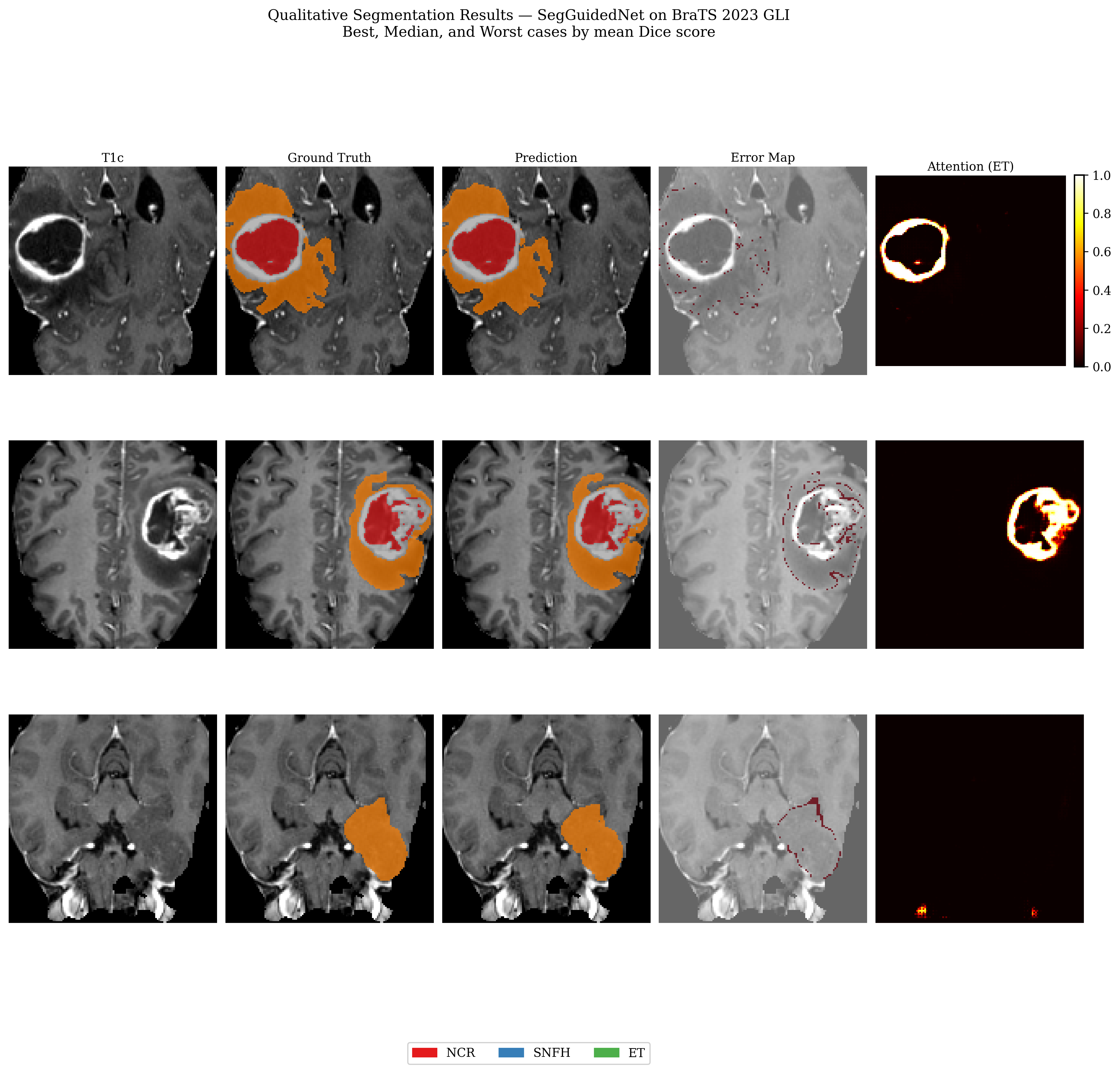}
    \caption{Qualitative results on BraTS\,2023 GLI best, median,
    and worst subjects (rows). Same column layout as
    Fig~\ref{fig:qualitative_2021}.}
    \label{fig:qualitative_2023}
\end{figure}

Figs~\ref{fig:qualitative_2021} and~\ref{fig:qualitative_2023}
present qualitative segmentation results for the best, median, and
worst-performing test subjects on BraTS\,2021 and BraTS\,2023 GLI
respectively. Across both datasets, the best and median cases
demonstrate highly accurate delineation of all three sub-regions
with minimal error, while the worst cases consistently reveal
under-segmentation of small or diffuse NCR regions the dominant
failure mode. The ET attention maps produced by the SegAttentionGate
closely correspond to ground-truth ET boundaries across both
benchmarks, confirming that the auxiliary supervision signal
generalises across dataset versions. Per-sub-region attention maps
and failure case analyses are provided in below
Figs.
\FloatBarrier 

\FloatBarrier 

\begin{figure}[!htbp]
    \centering
    \includegraphics[width=\textwidth]{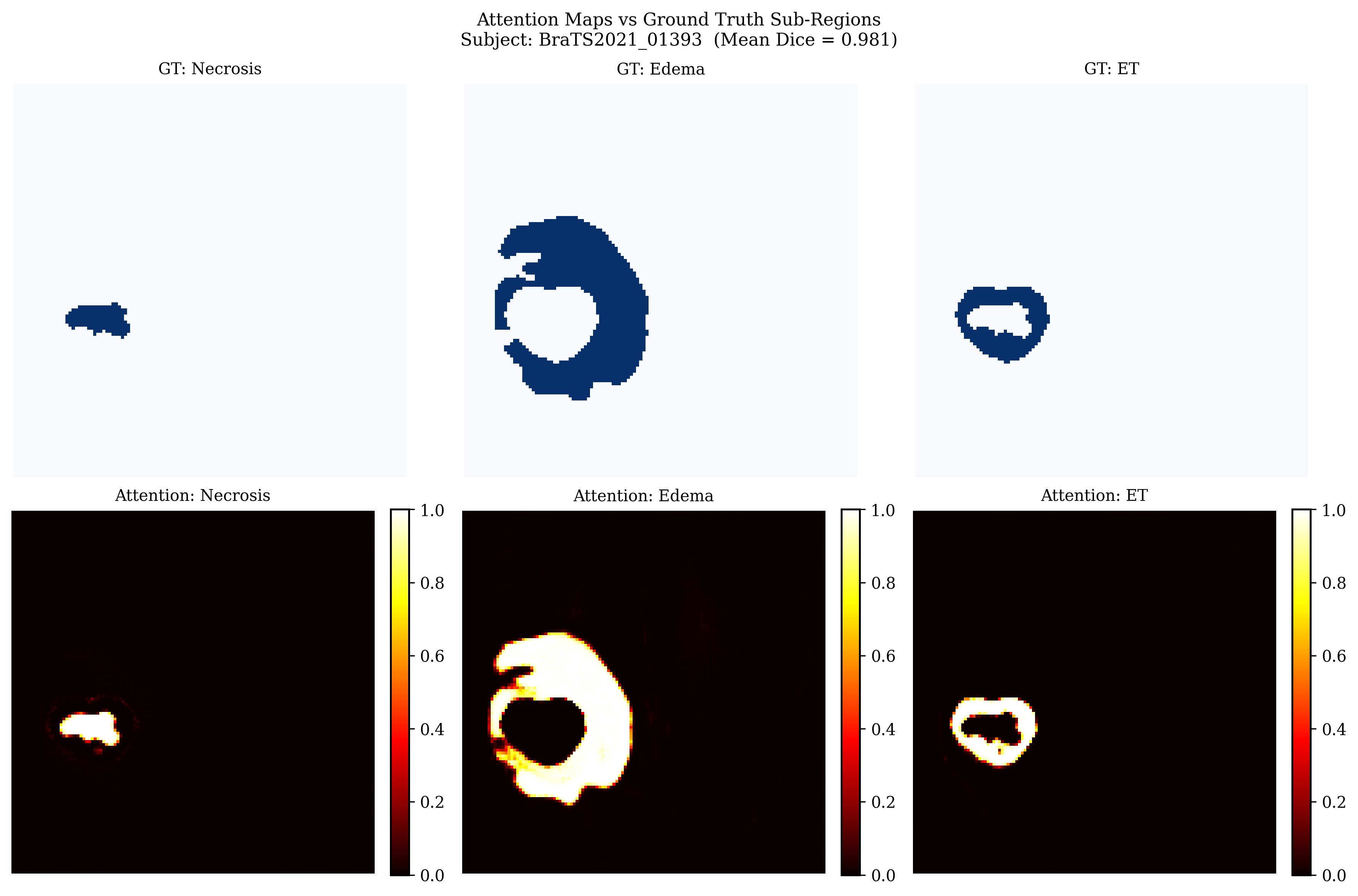}
    \caption{SegAttentionGate maps vs ground-truth sub-region masks
    for the best BraTS\,2021 subject. Top: GT binary masks.
    Bottom: learned attention maps (hot colourmap, range $[0,1]$).}
    \label{fig:attention_2021}
\end{figure}

As illustrated in Fig.~\ref{fig:attention_2021}, the proposed SegAttentionGate effectively concentrates on clinically significant tumor sub-regions by producing high-response attention areas that closely correspond to the ground-truth masks. The learned attention maps demonstrate the model’s ability to capture discriminative spatial representations, thereby enhancing segmentation accuracy and boundary localization for heterogeneous tumor structures in the BraTS 2021 dataset.
\FloatBarrier 

\begin{figure}[!htbp]
    \centering
    \includegraphics[width=\textwidth]{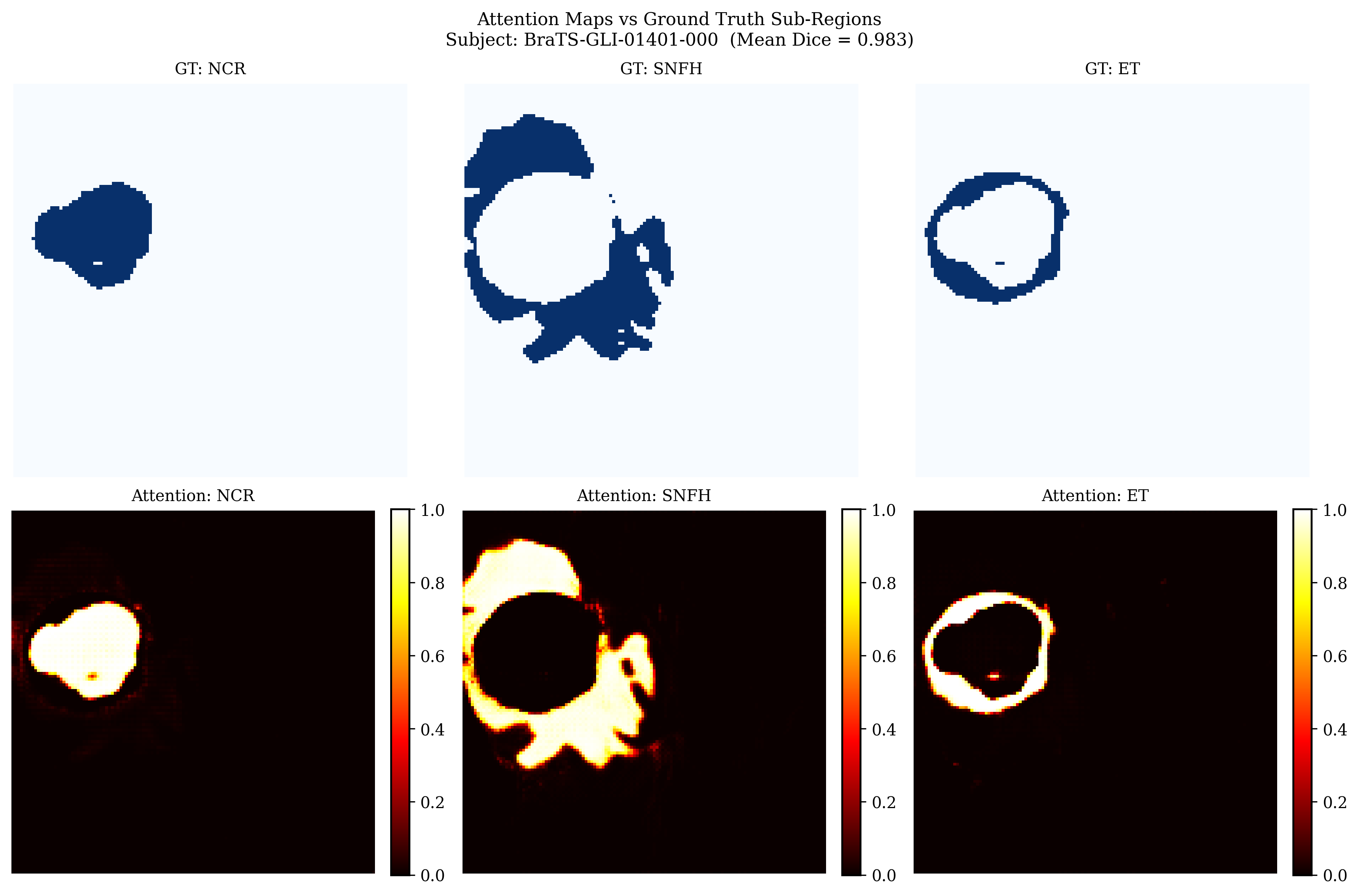}
    \caption{SegAttentionGate maps vs ground-truth sub-region masks
    for the best BraTS\,2023 GLI subject.}
    \label{fig:attention_2023}
\end{figure}

Fig~\ref{fig:attention_2023} illustrates the visualization of SegAttentionGate maps alongside the corresponding ground-truth tumor sub-region masks for the best-performing BraTS 2023 GLI subject. The attention maps demonstrate that the proposed model effectively focuses on clinically relevant tumor regions, highlighting enhanced localization capability and improved segmentation consistency across different tumor sub-regions.

\FloatBarrier 

\begin{figure}[!htbp]
    \centering
    \includegraphics[width=\textwidth]{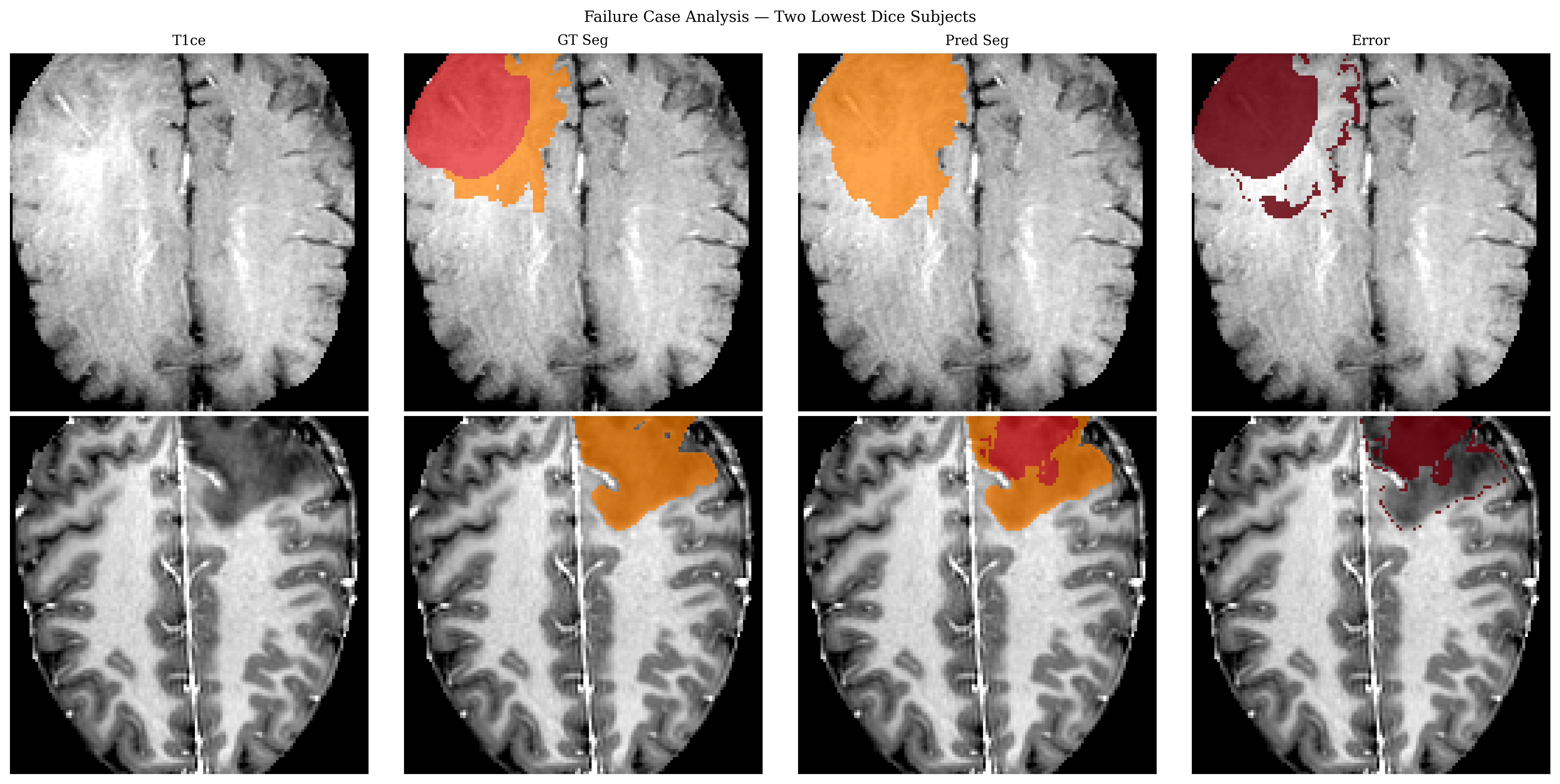}
    \caption{Failure cases on BraTS\,2021 — two lowest-Dice subjects.
    Columns: T1ce, GT, prediction, error map.}
    \label{fig:failure_2021}
\end{figure}

Fig~\ref{fig:failure_2021} illustrates the failure analysis of the proposed model on the two lowest-Dice subjects from the BraTS 2021 dataset. The results show that most segmentation errors occur around irregular tumor boundaries and small enhancing tumor regions, leading to partial misclassification between predicted and ground-truth masks.
\FloatBarrier 

\begin{figure}[!htbp]
    \centering
    \includegraphics[width=\textwidth]{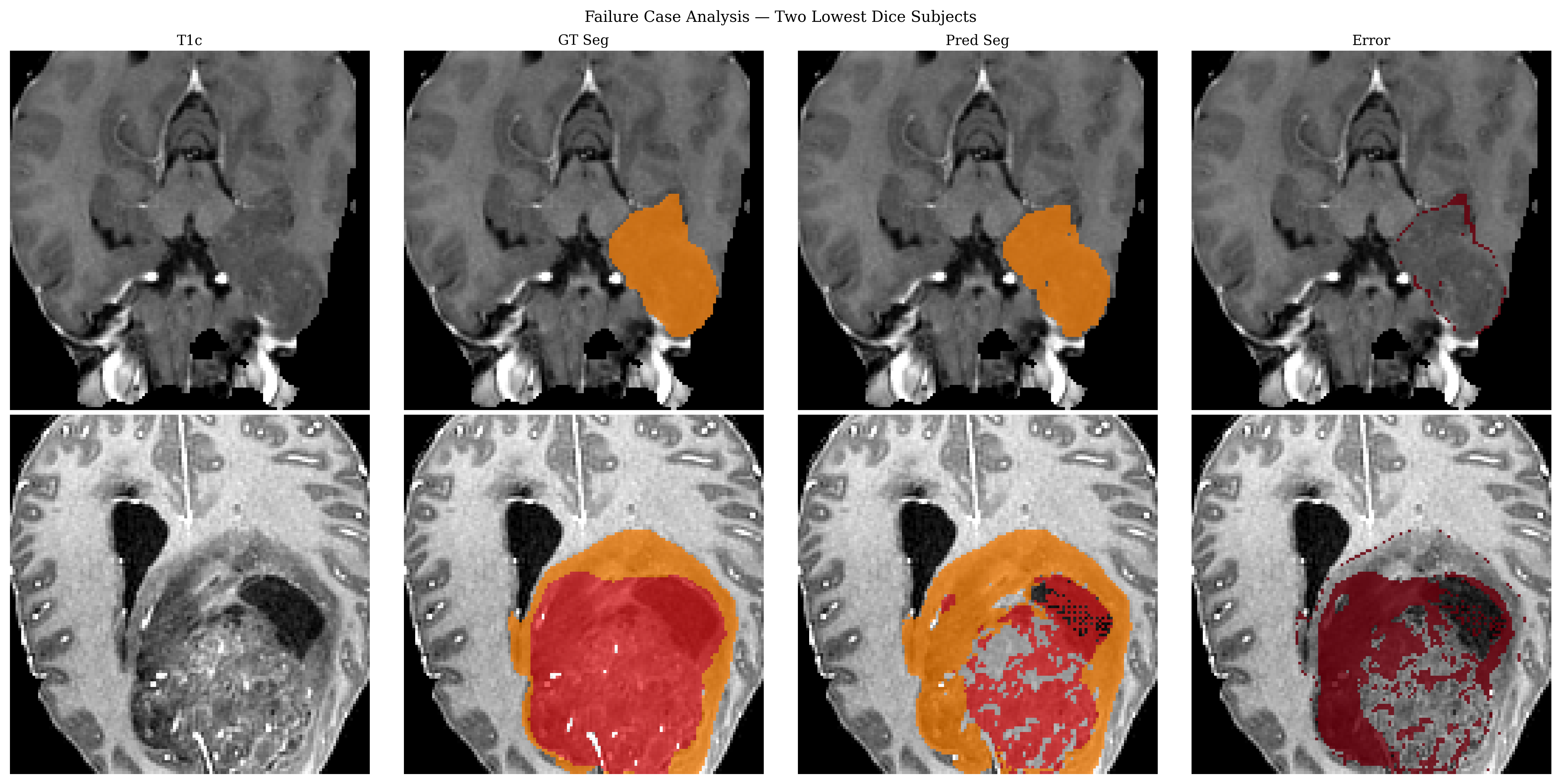}
    \caption{Failure cases on BraTS\,2023 GLI — two lowest-Dice subjects.}
    \label{fig:failure_2023}
\end{figure}

Fig~\ref{fig:failure_2023} presents the failure analysis on the BraTS 2023 GLI dataset by illustrating the two lowest-Dice subjects obtained by the proposed model. The results highlight challenging cases with complex tumor boundaries, low contrast regions, and irregular lesion structures, which negatively affected the segmentation accuracy and led to higher prediction errors.

\FloatBarrier 

\section{Discussion}\label{sec:discussion}

SegGuidedNet achieves strong brain tumour segmentation performance
across both benchmarks evaluated in this study. On BraTS\,2021, the
model attains a mean Dice of 0.905 (ET\,=\,0.873,
TC\,=\,0.906, WT\,=\,0.935) with HD95 below 4.0\,mm, while on
BraTS\,2023 GLI it achieves a mean Dice of \,0.897
(ET\,=\,0.859, TC\,=\,0.902, WT\,=\,0.931) with HD95 below
5.2\,mm both on held-out test sets of 251 subjects. The
consistency of these results across two independent benchmark
editions strengthens confidence in the generalisability of the
proposed approach beyond a single dataset. Critically, these
figs are obtained with a single model trained without
ensembling, test-time augmentation, or post-processing
heuristics conditions under which most competing methods
report their results. Compared to nnU-Net~\cite{isensee2021nnunet}
and HNF-Netv2~\cite{jia2021hnf}, SegGuidedNet surpasses both
across all three regions on BraTS\,2021, and produces equivalent
improvements on BraTS\,2023 GLI. While Swin
UNETR~\cite{hatamizadeh2021swin} retains a margin of 2--4 Dice
points, it does so with a 10-model ensemble; the gap narrows
considerably when inference cost is accounted for, making
SegGuidedNet a more practical choice for resource-constrained
clinical deployment.

The contribution of the SegAttentionGate is evident in both
quantitative and qualitative evaluations across both datasets.
The attention map visualisations reveal close spatial
correspondence between the learned sub-region maps and
ground-truth annotations particularly for ET, the most
clinically critical region for treatment response monitoring.
Notably, this interpretability arises directly from the
training objective and generalises across BraTS\,2021 and
BraTS\,2023 GLI without any architectural modification,
confirming that the auxiliary supervision signal captures
genuine sub-region discriminability rather than
dataset-specific artefacts. The strong TC performance
(0.906 and 0.902 respectively) further supports this
interpretation, as TC is a compound region combining NCR
and ET two sub-regions with overlapping T1ce signal
whose disambiguation directly benefits from explicit
sub-region supervision.

The dominant failure mode, consistent across both benchmarks,
is under-segmentation of small or diffuse NCR regions,
reflected in the higher Dice variance for that sub-region
(std\,=\,0.296 on BraTS\,2021; std\,=\,0.267 on
BraTS\,2023 GLI). This is well-documented across BraTS
methods, where NCR is the smallest and most visually
ambiguous sub-region~\cite{menze2015multimodal}. Near zero
ET Dice outliers on both datasets correspond to cases with
clinically absent enhancing tumour a known characteristic
of the BraTS cohort rather than systematic model failure.
Future work will explore sliding-window inference for
peripheral tumours, uncertainty quantification via Monte
Carlo dropout, and cross-dataset transfer to BraTS
metastasis and paediatric tumour tracks.

\section{Conclusion}\label{sec:conclusion}

We presented SegGuidedNet, a 3-D residual encoder--decoder
network that introduces the SegAttentionGate a lightweight
auxiliary module that explicitly supervises the decoder to
produce spatially discriminative attention maps for each
tumour sub-region via a dedicated binary cross-entropy loss,
adding less than 0.2\% parameter overhead. Evaluated
independently on BraTS\,2021 and BraTS\,2023 GLI across
251 held-out subjects each, SegGuidedNet achieves mean Dice
of 0.905 and 0.897 respectively,
outperforming ensemble-based nnU-Net and HNF-Netv2 as a
single model and approaching Swin UNETR a 10-model
ensemble within 2--4 Dice points at a fraction of the
inference cost. Beyond segmentation accuracy, the
SegAttentionGate provides interpretable sub-region attention
maps at no additional inference cost, with consistent spatial
fidelity across both benchmark editions. These results
demonstrate that explicit sub-region attention supervision
is an effective, efficient, and generalisable inductive bias
for brain tumour segmentation, offering a practical path
toward accurate and interpretable single-model performance
suitable for clinical neuro-oncology workflows.

\section*{Declarations}
\begin{itemize}
\item \textbf{Ethics approval and consent to participate}: Not applicable.
\item \textbf{Funding.} No funding
\item \textbf{Declaration of competing interest.} The authors declare that they have no known competing financial interests or personal relationships that could have appeared to influence the work reported in this paper.
\item \textbf{Consent for publication} Not applicable
\item \textbf{Data availability.} Available from corresponding author upon request.
\bmhead{Code availability}
PyTorch implementation, model weights, and documentation will be available on GitHub upon publication.
\item \textbf{CRediT authorship contribution statement.} Hasaan Maqsood \& Saif Ur Rehman Khan: Conceptualization, Data curation, Methodology, Software, Validation, Writing original draft \& Formal analysis. Muhammed Nabeel Asim, Sebastian Vollmer \& Andreas Dengel: Conceptualization, Funding acquisition, Review.
\end{itemize}

\section*{Abbreviations}\label{sec:abbrev}
Table~\ref{tab:abbreviations} lists the key abbreviations
used throughout this paper.

\begin{table}[htbp]
\centering
\caption{Abbreviations and Definitions}
\label{tab:abbreviations}
\begin{tabular}{lp{0.7\linewidth}}
\toprule
\textbf{Abbreviation} & \textbf{Definition} \\
\midrule
\multicolumn{2}{l}{\textbf{Clinical \& Medical Imaging}} \\
\midrule
MRI   & Magnetic Resonance Imaging \\
mpMRI & Multi-Parametric Magnetic Resonance Imaging \\
T1ce  & Contrast-Enhanced T1-Weighted MRI Sequence \\
FLAIR & Fluid-Attenuated Inversion Recovery \\
NCR   & Necrotic Tumour Core \\
ED    & Peritumoral Oedematous Tissue \\
ET    & Enhancing Tumour \\
TC    & Tumour Core (NCR $+$ ET) \\
WT    & Whole Tumour (NCR $+$ ED $+$ ET) \\
\midrule
\multicolumn{2}{l}{\textbf{Deep Learning}} \\
\midrule
CNN   & Convolutional Neural Network \\
IN    & Instance Normalisation \\
AMP   & Automatic Mixed Precision \\
BCE   & Binary Cross-Entropy Loss \\
AdamW & Adam Optimiser with Decoupled Weight Decay \\
\midrule
\multicolumn{2}{l}{\textbf{Evaluation Metrics}} \\
\midrule
DSC   & Dice Similarity Coefficient \\
HD95  & 95th-Percentile Hausdorff Distance \\
Sens. & Sensitivity \\
Spec. & Specificity \\
\bottomrule
\end{tabular}
\end{table}

\newpage
\bibliography{sn-bibliography}

\end{document}